\documentclass[letterpaper, 10pt, conference]{ieeeconf}

\IEEEoverridecommandlockouts
\usepackage[utf8]{inputenc}

\usepackage{times}

\usepackage{multirow}
\usepackage{xspace}
\usepackage{comment}
\usepackage{graphicx}
\usepackage{xcolor}
\definecolor{gg}{RGB}{215, 25, 28}

\usepackage{amsmath}
\usepackage{amssymb}
\usepackage{amsthm}
\usepackage{ctable}
\usepackage[flushleft]{threeparttable}
\usepackage{textcomp}

\newtheoremstyle{main}
                {1em}                                                % space above
                {1em}                                                % space below
                {\normalfont}                                        % bodyfont
                {0pt}                                                % indent
                {\scshape}                                           % head font
                {\\*}                                                % head punctuation
                {2pt}                                                % head space
                {\thmname{#1}\thmnumber{ #2}: \thmnote{\itshape #3}} % head spec

\theoremstyle{main}

\usepackage[linesnumbered,ruled,noend]{algorithm2e}

\usepackage[
  activate   = {true},
  protrusion = true,
  expansion  = true,
  kerning    = true,
  spacing    = true,
  tracking   = false,
  auto       = true,
  selected   = true,
  factor     = 1000,
  stretch    = 10,
  shrink     = 25,
]{microtype}

\makeatletter
\let\NAT@parse\undefined
\makeatother
\usepackage[pdfa,colorlinks,bookmarksopen,bookmarksnumbered,allcolors=gg]{hyperref}
\usepackage{cite}
\newcommand{\spark}{\textsc{spark}\xspace}

\newcommand{\sparkthree}{\textsc{spark}\xspace}
\newcommand{\flame}{\textsc{flame}\xspace}
\newcommand{\thunder}{\textsc{thunder}\xspace}
\newcommand{\cvae}{\textsc{cvae}\xspace}
\newcommand{\gmm}{\textsc{gmm}\xspace}
\newcommand{\ompl}{\textsc{ompl}\xspace}
\newcommand{\vicon}{\textsc{Vicon}\xspace}
\newcommand{\C}{\ensuremath{\mathcal{C}}\xspace}

\newcommand{\DB}{\ensuremath{\mathcal{B}}\xspace}
\newcommand{\W}{\ensuremath{\mathcal{W}}\xspace}

\newcommand{\cspace}{\mbox{\ensuremath{\mathcal{C}}-space}\xspace}
\newcommand{\regions}{\mbox{``challenging regions''}\xspace}
\newcommand{\regionsp}{\mbox{``challenging regions.''}\xspace}
\newcommand{\regionsc}{\mbox{``challenging regions,''}\xspace}
\newcommand{\norm}[1]{\left\lVert#1\right\rVert}
\newcommand{\se}{\ensuremath{\textsc{se}(\oldstylenums{3})}\xspace}
\newcommand{\distse}[2]{\ensuremath{d_{\se}\left(#1, #2\right)}\xspace}
\newcommand{\inprod}[1]{\left\langle#1\right\rangle}
\newcommand{\rrtc}{\textsc{rrt}-Connect\xspace}
\newcommand{\prm}{\textsc{prm}\xspace}

\newcommand{\threed}{\oldstylenums{3}\textsc{d}\xspace}
\newcommand{\twod}{\oldstylenums{2}\textsc{d}\xspace}
\newcommand{\dof}{\textsc{dof}\xspace}

\newcommand{\lw}{\ensuremath{\ell}\xspace}
\newcommand{\LW}{\ensuremath{\mathcal{L}}\xspace}

\newcommand{\wo}{\ensuremath{\omega}\xspace}

\newcommand{\gnat}{\textsc{gnat}\xspace}

\newcommand{\probci}[3]{\ensuremath{P_{#3}\left(#1\mid#2\right)}\xspace}
\newcommand{\probc}[2]{\ensuremath{P\left(#1\mid#2\right)}\xspace}

\newcommand{\obs}{\text{obs}\xspace}

\newcommand{\pv}{\ensuremath{p}\xspace}

\newcommand{\drad}{\ensuremath{d_\text{radius}}\xspace}

\title{\LARGE \bf
Learning Sampling Distributions Using Local \threed  Workspace Decompositions for Motion Planning in High Dimensions}

\author{Constantinos Chamzas, Zachary Kingston, Carlos Quintero-Pe\~na,\\ Anshumali Shrivastava, and Lydia E. Kavraki  % <-this % stops a space
  \thanks{CC is supported by NSF-GRFP 1842494 and ZK is supported by NSTRF 80NSSC17K0163. This work is also was supported in part by NSF 1718478, and Rice University Funds.}% <-this % stops a space
  \thanks{
      CC, ZK, CQP, AS and LEK are with the  
      Department of Computer Science, % 
      Rice University, %
      Houston, TX, USA  %
    {\tt \footnotesize \{chamzas,zak,carlosq,anshumali,kavraki\}@rice.edu}
  }
}

\begin{document}

\maketitle
\thispagestyle{empty}
\pagestyle{empty}

\begin{abstract}
  Earlier work has shown that reusing experience from prior motion planning problems can improve the efficiency of similar, future motion planning queries.
However, for robots with many degrees-of-freedom, these methods exhibit poor generalization across different environments and often require large datasets that are impractical to gather.
We present \sparkthree and \flame, two experience-based frameworks for sampling-based planning applicable to complex manipulators in \threed environments.
Both combine samplers associated with features from a workspace decomposition into a global biased sampling distribution.
\sparkthree decomposes the environment based on exact geometry while \flame is more general, and uses an octree-based decomposition obtained from sensor data.
We demonstrate the effectiveness of \sparkthree and \flame on a real and simulated Fetch robot tasked with challenging pick-and-place manipulation problems.
Our approaches can be trained incrementally and significantly improve performance with only a handful of examples, generalizing better over diverse tasks and environments as compared to prior approaches.

\end{abstract}

\section{Introduction}

\label{sec:introduction}

For many high-dimensional systems, sampling-based methods have proven very successful at efficient motion planning~\cite{Choset2005}.
However, as noted by~\cite{Hsu2003}, they are sensitive to \regions, such as narrow passages in the search space, which have a low probability of being sampled.
The existence of such \regions hinders the performance of sampling-based planners.

Many methods have improved the performance of sampling-based planners by using \emph{experience} from prior motion planning problems, e.g.,~\cite{Coleman2015, Ichter2018}.
In particular, experience can be used to learn how to explore the \regions of the search space.
However, for high dimensional robots, learning-based methods typically face difficulties in generalizing even between similar environments~\cite{Chamzas2019, Kumar2019, Tang2019}.

This work presents two experience-based planning frameworks---one using exact geometry (\sparkthree) and one using sensor data (\flame)---for high \dof robots in \threed environments.
Both frameworks build on the decomposition paradigm introduced by~\cite{Chamzas2019}. That work introduced the idea of \emph{local primitives}, which capture important local workspace features.
Each primitive is associated with a \emph{local sampler}, which produces samples in \regions created by the primitive, and are stored in an experience database.
The database is updated offline by adding experiences from past planning problems in an incremental manner.
Given a planning query, similar local primitives to those in the current workspace are retrieved; their local samplers are combined and used by a sampling-based planner.
In such a framework several questions arise on the definition and use of local primitives.
\cite{Chamzas2019} required \emph{a priori} knowledge of local primitives and was designed for \twod environments with circular obstacles.

The contributions of this paper are the following.
We present \sparkthree which scales to \threed environments when exact geometric information is available and \flame which relies only on sensor information.
In order to develop these frameworks, we revisit and enrich the framework of \cite{Chamzas2019} by introducing novel local primitives, distance functions, and utilizing efficient retrieval structures.
Furthermore, we propose a new \emph{criticality test} that enables incremental training of \spark and \flame from novel scenes. 
Through the synergistic combination of the introduced components, we achieve real-time retrieval and training of both frameworks. Finally we validate our methods on a suite of simulated and real robot experiments with the Fetch robot, an 8 degree-of-freedom (\dof) manipulator.
Compared to prior work, our methods exhibit significantly better performance over diverse environments and tasks with relatively few examples. The datasets and the implementation of \spark and \flame are open-sourced \footnote{\label{fn:website}{\mbox{\url{https://github.com/KavrakiLab/pyre}}}}.     

\begin{figure}
  \centering
	 \includegraphics[width=\linewidth]{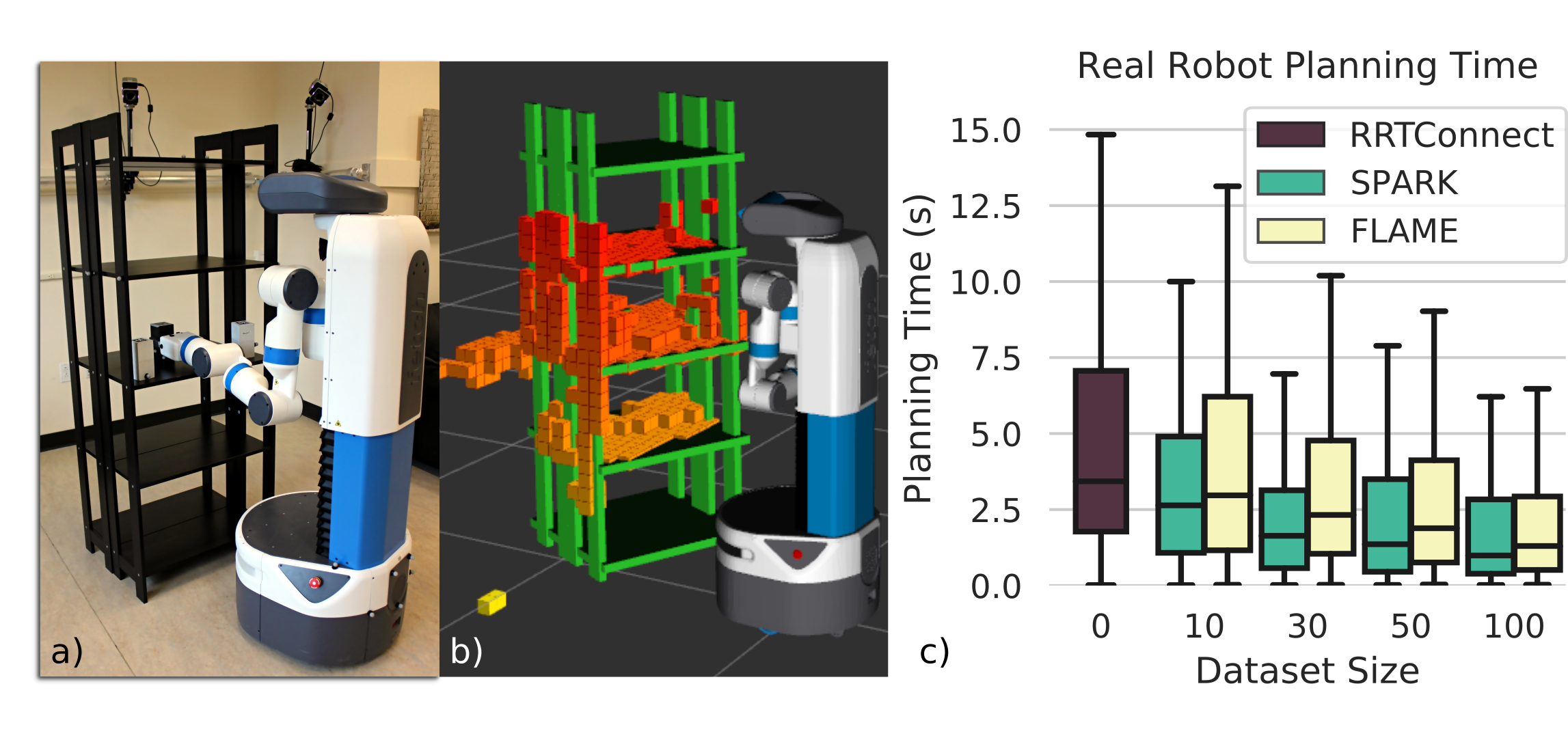}
    \vspace{-2em}
   \caption{ The proposed sampling-biasing frameworks applied in a real challenging motion planning problem.
\textbf{a)} The 8 \dof Fetch robot must reach into a shelf to grasp one of the randomly positioned blocks.
Between problems, the base pose varies by up $\pm 90^\circ$ in angular position and up to $\pm10$ cm along the X- and Y-coordinates.
\textbf{b)} A \vicon camera system was used to gather object poses for \sparkthree.
The Fetch's depth camera was used to build an octomap~\cite{Hornung2013} for \flame which can retrieve relevant experience even from partial noisy octomaps.
\textbf{c)} Both improve \rrtc even with few given examples.}%
	\label{fig:real}
\end{figure}

\section{Problem Statement}
\label{sec:formulation}

Consider a robot in a workspace \W.
A robot configuration $x$ is a point in the configuration space (\cspace), $x \in \C$.
Obstacles in workspace induce \cspace obstacles where the robot is in collision, $X_\obs \subset \C$, which must be avoided.
We are interested in finding a collision-free \emph{path}, from start to goal as efficiently as possible.

Sampling-based planners are widely used for motion planning~\cite{Choset2005, LaValle2006}.
These planners randomly sample to explore the \cspace.
However, the performance of sampling-based planners is degraded when \regions exist in the \cspace~\cite{Hsu2003}.
These regions, introduced by $X_\obs$, create difficult-to-explore areas in the \cspace due to their ``low visibility'' relative to the entire \cspace.
It is understood~\cite{Hsu1999, Hsu2006} that biased sampling that targets such regions can significantly improve performance.
Thus, the problem we consider is how to learn to \emph{bias sampling} towards \regions using workspace information and prior experiences. 

\section{Related Work}
\label{sec:related}

One way to bias sampling is with rejection sampling, which draws uniform samples but only accepts them if a geometric test is passed, e.g., bridge sampling~\cite{Hsu2003} or Gaussian sampling~\cite{Boor1999}.
However, rejection sampling is inefficient, possibly drawing many samples before a valid one is found.
Some approaches attempt to directly sample the \regionsc such as medial axis
sampling~\cite{Lien2003} or free-space dilation~\cite{Hsu1998}, but these distributions become difficult to compute as the state space increases in dimension.

Rather than using fixed distributions, many methods \cite{Morales2004,Tapia2009} \emph{adapt} sampling online based on collision information, such as utility-guided sampling~\cite{Burns2005}, toggle-\prm~\cite{Denny2013} or hybrid-\prm\cite{Hsu2005}.
However, these methods do not transfer knowledge between planning queries.
Our methods bias based on prior motion plans, and thus is complementary to adaptive sampling methods.

Another category of techniques improves planning by learning problem invariants. Examples include using sparse roadmaps~\cite{Coleman2015}, informed lazy evaluations~\cite{Bhardwaj2019}, path databases~\cite{Berenson2012, Phillips2012,Lembono2020Memory}, or biased distributions~\cite{Lehner2018}. However, these methods do not use workspace information. Thus, changes in the environment force these methods to repair now invalidated information, reducing performance.

On the other hand, many methods do use workspace features to guide sampling---typically, exact~\cite{Kurniawati2004, Kurniawati2008} or inexact~\cite{Zucker2008} workspace decompositions are used.
For example, the authors of~\cite{Brock2001} use balls to decompose the workspace.
However, to generate samples in \regionsc these approaches require a mapping from workspace to \cspace, reliant on the robot's inverse kinematics.
Thus, these methods are typically limited to free-flying or low-dimensional robots, as manipulator kinematics can become complex.
The proposed methods also use workspace decompositions (both an exact and inexact one), but instead \emph{learn} the inverse mapping from workspace to \cspace.

Many recent techniques use deep learning to learn the mapping from workspace to \cspace, such as using neural networks to bias sampling~\cite{Ichter2018, Zhang2018, Kumar2019, Chen2020} or to guide planners~\cite{Qureshi2019,Tamar2019, Terasawa20203d}).
However, due to the inherent discontinuity of motion planning~\cite{Hauser2018, Farber2003}, this mapping is generally discontinuous, limiting the applicability of these methods~\cite{Tang2019, Chamzas2019, Merkt2020}.
That is, small changes in the workspace result in substantial (possibly discontinuous) changes in \cspace.  
Thus, these approaches usually apply only to manipulators with low workspace variance~\cite{Ichter2018,Zhang2018,Qureshi2019,Chen2020}.

On the other hand, a category of methods that addresses the discontinuity problem, are mixture model methods and similarity-based methods.
Mixture model methods~\cite{Tang2019, Finney2007} learn different functions for different discontinuous regions but are limited by the fixed number of mixtures.
Similarity-based methods~\cite{Lien2009, Jetchev2013, Chamzas2019}, which are most similar to ours, are usually non-parametric techniques that retrieve relevant information (e.g., \regions representations) based on workspace similarity.
Retrieval is by design discontinuity-sensitive since multiple different \regions can be retrieved for similar workspaces.
However,~\cite{Lien2009} is limited to free flying robots~\cite{Jetchev2013} addresses trajectory optimization-based problems rather than sampling and~\cite{Chamzas2019} is limited to \twod planar environments with circular objects.
Our work shares the retrieval idea with these methods but applies to general \threed workspaces with realistic robots.

\section{Methodology}
\label{sec:methodology}

We introduce two experience-based planning frameworks for arbitrary \threed environments and manipulators, \sparkthree and \flame.
\sparkthree requires an exact representation of the environment's geometry, while \flame only uses sensor information represented as an octomap~\cite{Hornung2013}.

Consider the Fetch robot reaching into a deep shelf, as shown in \autoref{fig:real}.
The arrangement of the workspace obstacles constricts the manipulator's movement, making the motion planning problem challenging.
Both frameworks decompose the workspace into \emph{local primitives} to capture local workspace features that create \regions in \cspace, e.g., the shelf's sides that create a narrow region for reaching.
Our frameworks learn by using a \emph{criticality test}, which associates ``critical'' configurations from every successful motion plan to relevant local primitives in the current workspace.
For example, the criticality test should capture that configurations of the Fetch's arm deep in the shelf, should be associated with the sides of the shelf.
These configurations are used to create a \emph{local sampler}---a biased sampling distribution focused on \regions created by the associated local primitive.
During learning, these primitive-sampler pairs are stored in a spatial database (\autoref{alg:1}).
When inferring for a new problem, relevant local primitive-sampler pairs are retrieved from the database; local samplers are then combined into a global biased sampler (\autoref{alg:2}).

\begin{algorithm}
  \caption{Learning \label{alg:1}}

  \SetKwInOut{Input}{Input}
  \SetKwInOut{Output}{Output}

  \Input{Workspace $\W$, Path \pv, Database \DB}
  \Output{Database \DB}
    Decompose $\W$ to $\LW \gets \{\lw_{1}, \ldots, \lw_{N}\}$ \label{alg:1:decompose} \\
    \ForEach{$\lw_{j} \in \LW$}{
      \textsc{critical}  $\gets\emptyset$ if $\lw_{j} \not\in \DB$ else $X_j$\\
      \ForEach{$x \in \pv$}{ \label{alg:1:critical}
        \If{\textsc{isCritical}($x$, $\lw_i$)}
           {
             \textsc{critical} $\gets$ \textsc{critical} $\cup~x$
           }
      }
      Generate \probci{x}{\lw_j}{j} from \textsc{critical}\\ \label{alg:1:generate}
      Insert $\langle \lw_j,\probci{x}{\lw_j}{j} \rangle$ in \DB \label{alg:1:store}
  }
  \Return \DB
\end{algorithm}

\subsection{Overall Framework}
\label{subsec:framework}
Both \sparkthree and \flame define their own decomposition of the workspace into local primitives, a criticality test for these primitives, and a metric to compare local primitives.
The database that stores local primitives, their associated local samplers, and the synthesis of the global sampler is the same in \sparkthree and \flame.

\subsubsection{Local Primitives}
Local primitives are workspace features that come from any valid decomposition of the workspace.
Each local primitive attempts to capture the workspace features that create \regions in the \cspace.
The definition of local primitives is a key component of the algorithm.
The local primitives for \sparkthree and \flame are given in their respective sections, and are shown in \autoref{fig:pairs} and \autoref{fig:octo}.
We denote local primitives with \lw.
From a given workspace \W, $N$ local primitives are created from a workspace decomposition such that $\bigcup_{i=1}^{N} \lw_i\subseteq \W$ (\autoref{alg:1:decompose} in \autoref{alg:1}, \autoref{alg:2:decompose} in \autoref{alg:2}).

\subsubsection{Local Samplers}
Each local primitive $\lw_i$ is associated with a local sampler, which should produce samples in the \regions created by the associated local primitive.
A local sampler $\probci{x}{\lw_i}{i}$ is a generative distribution over a set of $M_i$ ``critical'' configurations $X_i = \{ x_{i1}, \dots, x_{iM_i} \}$ gathered from past experience.
Due to the potential complexity of this distribution and its natural multi-modality, we represent the local sampler as a Gaussian mixture model (\gmm), similar to~\cite{Lehner2018}.
Contrary to~\cite{Lehner2018}, we do not use expectation maximization to calculate the parameters of the \gmm.
Instead, we place one mixture for each configuration $x_{ij}$ in the local sampler $P_i$ with a fixed covariance $\sigma$ (\autoref{alg:1:generate} in \autoref{alg:1}):
\begin{equation*}
  \label{eq:1}
  \probci{x}{\lw_i}{i} = \frac{1}{M_i}\sum_{j=1}^{M_i} \mathcal{N}(x_{ij}, \sigma).
\end{equation*}
We choose uniforms weights---all mixtures are equiprobable.

\subsubsection{Criticality Test}
In \threed local primitives may or may not create \regionsc and finding if a configuration belongs to such a region is not straightforward. 
The authors of~\cite{Molina2019} chose regions with maximum path density while~\cite{Ichter2020} and~\cite{Kumar2019} identified \regions using graph properties of a roadmap.
However in our setting only finding these regions is not enough, since they must also correspond to a local primitive.
To overcome this problem, and depart from earlier work, we introduce a \emph{criticality test} (\textsc{isCritical}, \autoref{alg:1:critical} in \autoref{alg:1}), which associates key configurations from solution paths to relevant local primitives.
The intuition is that a configuration in a critical region will be close to a \cspace obstacle, which often means that some part of the robot will be close to a workspace obstacle.  
Both \sparkthree and \flame define their own criticality test, which enables learning incrementally without \emph{a priori} knowledge. 

\begin{algorithm}
  \caption{Inference \label{alg:2}}

  \SetKwInOut{Input}{Input}
  \SetKwInOut{Output}{Output}

  \Input{Query Workspace $\W$, Database \DB}
  \Output{Sampler \probc{x}{\W}}
  Decompose \W to $\LW \gets \{\lw_{1}, \ldots, \lw_{N}\}$ \label{alg:2:decompose} \\
  \ForEach{$\lw_{j} \in \LW$}{
    Retrieve all $\{ \probci{x}{\lw_i}{i} \mid d(\lw_i, \lw_j) \leq \drad \}$ \label{alg:2:retrieve}
  }
  Synthesize \probc{x}{\W} from all $\probci{x}{\lw_i}{i}$ \\\label{alg:2:synth}
  \Return \probc{x}{\W}
\end{algorithm}

\subsubsection{Distance Function}
Given a new motion planning problem, the workspace is decomposed into a set of local primitives.
Based on a \emph{distance} metric over local primitives $d(\lw_i, \lw_j)$, we search for similar local primitives in the existing database $d(\lw_i, \lw_j) \leq \drad$ and retrieve their associated local samplers (\autoref{alg:2:retrieve} in \autoref{alg:2}).
Both \sparkthree and \flame define a distance metric over their local primitives.

Quick retrieval of relevant local primitives is essential to our method.
We use  \gnat~\cite{Brin1995}, a data structure with logarithmic scaling for nearest-neighbor retrieval in high-dimensional metric spaces.
\gnat also enables fast rejection of queries that do not exist within the database, which is useful as many local primitives are irrelevant e.g., outside the reachable workspace.

\subsubsection{Global Sampling}
Given a set of $K$ retrieved local samplers, they are combined in a composite \gmm (\autoref{alg:2:synth} in \autoref{alg:2}):
\begin{equation*}
  \probc{x}{\W} \approx \frac{1}{K} \sum_{i=1}^K  \probc{x}{\lw} = \frac{1}{K} \sum_{i=1}^K  \frac{1}{M_i} \sum_{j=1}^{M_i}  \mathcal{N}(x_{ij}, \sigma).
\end{equation*}
To avoid redundancy as local samplers may be the same, a local sampler is used only if its unique.
To retain probabilistic completeness, there is a chance $\lambda > 0$ to sample uniformly rather than from the \gmm.
If no local samplers are retrieved the biased sampler defaults to uniform sampling, i.e., $\lambda = 1$.

%%
%% SPARK 3
%%

\subsection{\sparkthree}
\label{subsec:spark3d}
\begin{figure*}

  \centering
  \includegraphics[width=0.95\textwidth]{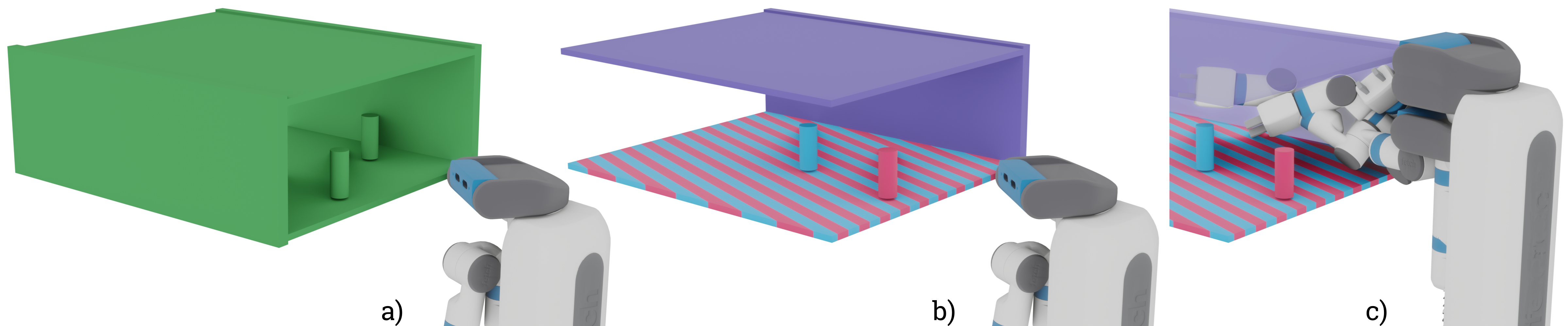}
  \caption{%
    An example of \spark, our experience-based planning framework that decomposes the workspace into pairs of known obstacles.
    \textbf{a)} The global workspace, composed of a set of boxes and cylinders.
    \textbf{b)} Three local primitives (different pairs are shown with blue, magenta, pink color) from the \spark decomposition.
    Note that the bottom of the shelf is shared between two primitives.
    \textbf{c)} A set of superimposed samples drawn from the local samplers.
  }
  \label{fig:pairs}
  \vspace{-1 em}
\end{figure*}

\subsubsection{Local Primitives}
In \sparkthree, local primitives are defined as pairs of box obstacles $\wo^a$ and $\wo^b$, which have the following parameters:
\begin{equation*}
\lw_{i} = \{\wo_i^a, \wo_i^b\} =  \{T_i^a, s_i^a, T_i^b, s_i^b\},
\end{equation*}
where $T, s$ respectively denote the pose and size---a \threed vector for the scale of each dimension---of each box, for a total of 10 parameters per box.
Non-box objects are approximated by bounding boxes.
Boxes that are too far from each other (see~\autoref{subsec:spark3d:dist}) are not considered valid local primitives.
An example of pairs of local primitives can be seen in~\autoref{fig:pairs}.

Intuitively, pairs of objects create \regions in the configuration space, as they constrict space in the workspace.
Note that one or many (e.g., 3) objects per local primitive are also viable options.
However, single objects are less discriminative than two and thus recall extraneous information.
In a similar vein, considering many objects combinatorially increases the number of primitives in a scene.
As we show in~\autoref{sec:experiments}, using pairs of objects provides good results in our tested scenarios.

\subsubsection{Criticality Test}

To check if a configuration is associated with a local primitive, we find the minimum distance between the robot at that configuration with the pair of obstacles.
For every shape in the robot's geometry, the distance to both objects in the local primitive is measured with the Gilbert-Johnson-Keerthi (GJK) algorithm~\cite{Gilbert1988}.
%The GJK algorithm is a standard algorithm for measuring the distance between convex objects.
If the minimum distance is smaller than a threshold $d_\text{clust}$, that configuration is related to that local primitive.

\subsubsection{Distance}
\label{subsec:spark3d:dist}
To measure the distance between poses in \se, we use~\cite{Kuffner2004}:
\begin{equation*}
  \label{eq:pose}
  \distse{T_i}{T_j} =  w_T \norm{t_i - t_j} + (1 - w_T)\left(1 - \inprod{q_i, q_j}^2\right),
\end{equation*}
where $t, q$ respectively denote translation and orientation (a quaternion) of a pose $T$, $w_T$ is a chosen weight, $\norm{\cdot}$ is the Euclidean norm and $\inprod{\cdot, \cdot}$ is the inner product.
To measure distance between boxes, we use a sum of metrics:
\begin{equation*}
  d_{\wo}(\wo_i, \wo_j) = w_s \distse{T_i}{T_j}  + (1 - w_s) \norm{s_i-s_j},
\end{equation*}
where $w_s$ is a chosen weight.
Local primitives for \spark are all the pairs of boxes $d_{\wo}(\wo_i, \wo_j)$ with distance less than $d_{\text{pairs}}$.
To compare local primitives, we use a metric that is the minimum distance between the boxes in each primitive:
\begin{align*}
  d(\lw_i, \lw_j) &= \min \begin{cases}
    d_{\wo}(\wo_i^a, \wo_j^a) + d_{\wo}(\wo_i^b, \wo_j^b)  \\
    d_{\wo}(\wo_i^a, \wo_j^b) + d_{\wo}(\wo_i^b, \wo_j^a)
    \end{cases}
\end{align*} 

%%
%% FLAME
%%

\subsection{\flame}
\label{subsec:flame}

\begin{figure*}
  \centering
  \includegraphics[width=0.95\textwidth]{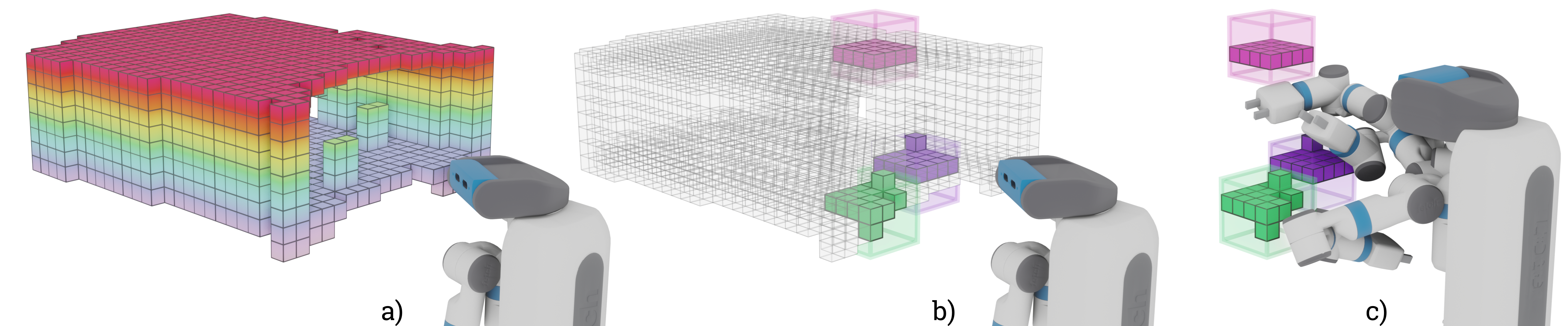}
  \caption{%
    An example of \flame, our experience-based planning framework that uses a decomposition of an octomap~\cite{Hornung2013} workspace gathered from sensing data.
    \textbf{a)} The global workspace as an octomap.
    \textbf{b)} Three local primitives (highlighted in color) from the \flame decomposition.
    Local primitives are $4 \times 4 \times 4$ occupancy grids derived from the global octomap.
    \textbf{c)} A set of superimposed samples drawn from the local samplers.
  }
  \label{fig:octo}
  \vspace{-1  em}
\end{figure*}

\subsubsection{Local Primitives}
In \flame, local primitives are defined as ``local'' occupancy grids, or \emph{octoboxes}, which are obtained from a global \emph{octree}.
An octree is a hierarchical volumetric tree that efficiently encodes obstacles in the workspace.
At each level, the octree contains voxels---descending a level of the tree, these voxels are split in half along each of its axes, each giving 8 subvoxels.
A voxel contains occupancy information at its resolution; a voxel is either occupied or free.
Voxels at higher levels can compact information from lower levels---if all subvoxels are either occupied or free, then the tree does not need to descend further.
\flame is useful for application on real robots as octrees~\cite{Hornung2013} are a common approach used to represent sensor data.

We define \emph{octoboxes} as occupancy grids that correspond to intermediate nodes of an octree.
Here, octoboxes contain the voxels of the two lowest levels of the octree, that is, 64 voxels as a $4 \times 4 \times 4$ \threed occupancy grid.

To decompose the workspace, an octree is built relative to the robot that captures the occupancy of the workspace.
Non-empty octoboxes are quickly generated by traversing the octree at a specific depth.
We also consider the pose $T$ of the center of the octobox in the global frame.
An example is shown in~\autoref{fig:octo}.
Local primitives in \flame are
$\lw_i = \{T_i, b_i\}$,
where $T$ is the pose of the octobox and $b$ is a $64$-bit occupancy grid vector, where each bit corresponds to an occupied ($1$) or free ($0$) voxel.

\subsubsection{Criticality Test}
A configuration $x_i$ is associated with a local primitive if the robot's geometry at configuration $x_i$ intersects the bounding box of the given octobox.
Note that a configuration may be associated with many local primitives.

\subsubsection{Distance}
To compare two octoboxes, we simply check if the two octoboxes are the same.
It may seem overly pessimistic to look for precise octobox matches, but our empirical results show that this simple metric yields good results.
One possible explanation is, since the criticality test assigns a configuration to many different local primitives, only one primitive needs to be recovered to recall the relevant configuration.
The distance metric is:
\begin{equation*}
  d(\lw_i, \lw_j) =
  \begin{cases}
     0  & b_i = b_j , T_i=T_j \\
    \infty & \text{otherwise} 
  \end{cases}
\end{equation*}

\section{Experiments}
\label{sec:experiments}

We evaluate the performance of \sparkthree and \flame in the following varied and realistic scenarios using a Fetch (8-\dof) robot manipulator.
We believe both frameworks can also be used with mobile/flying robots, but focus on manipulators since we consider them more challenging for learning methods. 
The ``Small Shelf'' environment (\autoref{fig:small}a) with 3 cylindrical objects, the ``Large Shelf'' environment with 7-9 cylindrical objects (\autoref{fig:large}a), the ``Real Shelf'' (\autoref{fig:real}) with 3 box objects and the "Box Table" with mutliple random objects (\autoref{fig:table}a).
In each environment, we vary both the pose of ``small'' objects (cylinders, boxes) and more importantly the pose of ``big'' objects(shelves, box, table) relative to the robot. 
The "big" objects variations of "Small shelf" are shown in \autoref{fig:small}a) and described in \autoref{fig:real}a), \autoref{fig:table}a), and \autoref{fig:large}a) for the other scenarios.
See the attached video for more details.

We consider these variations of environments highly challenging for learning-based motion planning frameworks for two reasons.       
First, the underlying motion planning problem is hard since it exhibits many \regionsp
Second, even small changes in the relative position of the objects can cause discontinuous changes in the configuration space obstacles, making it hard to learn functions that map from workspace to \cspace~\cite{Farber2003, Tang2019}.
For example, a small change in the relative angle of the shelf to the robot might require a solution that goes on the left side of the robot, versus the right side.
Such diversity is realistic, as mobile manipulators can approach objects from many different angles and positions.
 
We test two tasks: a ``pick'' and a ``place'' task.
In the ``pick'' task, the Fetch plans between its stow position and the farthest object on the shelf. 
In the ``place'' task in the ``Large Shelf'' environment, the Fetch starts rigidly grasping the object farthest back on the top or bottom shelf and places the object in the back of the middle shelf.
In the ``place'' task in the ``Box Table'' environment, the Fetch starts grasping the cube inside the box and place randomly on the table.

We implemented our framework using the Open Motion Planning Library (\ompl)~\cite{Sucan2012} and \emph{Robowflex}~\cite{robowflex}.
Note that \sparkthree and \flame both generate sampling distributions and must be used by a sampling-based planner.
For all experiments and frameworks, \rrtc~\cite{Kuffner2000} is used with default settings unless noted otherwise.
All the experiments were run on an Intel\textsuperscript{\tiny{\textregistered}} i7-6700 with four 4.00\textsc{GHz} cores and 32 \textsc{gb} of memory.
The hyperparameters for \spark(7) and \flame(3) are given in \autoref{tab:hyper} and were the same for all experiments.
These are reasonable defaults for our methods, found heuristically. The methods performance was empirically observed to be robust over a wide range of values. 
%We have empirically observed that \spark and \flame are robust to changes in these
%parameters and achieve good performance .

\begin{table}
	\centering
  \begin{threeparttable}
	  \caption{\sparkthree and \flame hyperparameters}
	  \label{tab:hyper}
    \begin{tabular}[t]{r|c|c}
      Parameter & Symbol & Value \\
      \specialrule{.3em}{.1em}{.1em}
      Sampling Variance       & $\sigma$                   & 0.2  \\\hline
		  Sampling Bias           & $\lambda$              & 0.5  \\\hline
		  Retrieval Radius        & $\drad$                & 0.4  \\\hline
		  \spark Clustering Distance  & $d_{\text{clust}}$     & 0.15 \\\hline
	      \spark Pair Distance        & $d_{\text{pairs}}$ & 0.2  \\\hline
		  \spark Pose Weight             & $w_{T}$           & 0.75 \\\hline
		  \spark Size Weight             & $w_{s}$           & 0.5 \\
      \specialrule{.3em}{.1em}{.1em}
	  \end{tabular}
  \end{threeparttable}
  % \vspace{-2em}
\end{table}%

\subsection{Generalization}
\label{subsec:var}
\begin{figure*}
    \vspace{1em}
  \centering
	\includegraphics[width=0.90\linewidth]{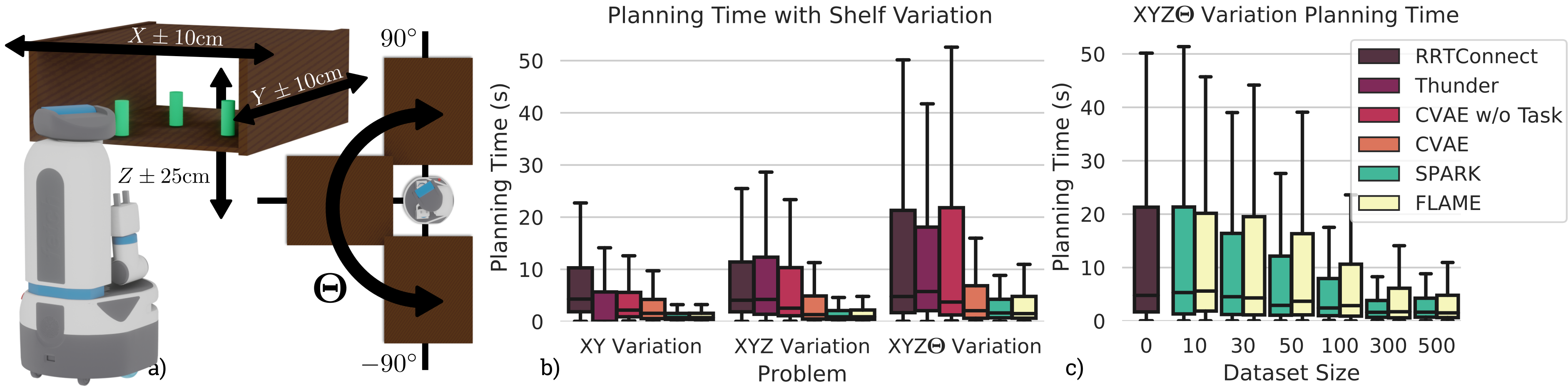}
  \caption{%
    \textbf{a)} The ``Small Shelf'' environment with a ``pick'' task.
    As shown, in this experiment, the orientation of the shelf relative to the robot varies up to $\pm 90^\circ$, in its X- and Y-coordinates by $\pm 10$cm, and by $\pm 25$cm in it Z-coordinate.
    \textbf{b)} Average planning time (including retrieval) versus the axes of variances. 500 training and 100 test environments were sampled for each, with a timeout of $60$ seconds.
    \textbf{c)} Average planning time (including retrieval) versus the size of the dataset, tested on 100 sampled environments with a timeout of $60$ seconds.
    %All methods use \rrtc as their sampling-based planner.
    %\sparkthree, and \flame outperform all methods when there is problem variance.
    %\thunder succeeds when the variance is low due to capturing the problem invariants well with its sparse roadmap but falls back to uniform sampling as problem variation grows.
    %\cvae greatly benefits from start and goal information, while our methods do not require this information.
    %\sparkthree and \flame improve quickly given a few examples.
  }
	\label{fig:small}

    \vspace{-0.5em}
\end{figure*}

\autoref{fig:small}b presents timing results for a ``pick'' task in ``Small Shelf'' over three types of variation, $XY$, $XYZ$, and $XYZ\Theta$ as shown in \autoref{fig:small}a.
We compare against \rrtc, an experience-based approach that uses a path database (\thunder)~\cite{Coleman2015}, and a conditional variational autoencoder framework (\cvae)~\cite{Ichter2018}.
\thunder stores previous paths as a sparse roadmap---new problems are solved by searching the roadmap for a valid path; if no path is found (possibly due to changes in the environment), a planner is used to repair the roadmap or find a new path from scratch.
The \ompl version of \thunder was used within \emph{Robowflex}.

\cvae learns a mapping from workspace to ``interesting samples'' in \cspace.
To infer new samples the latent space of the \cvae can be sampled and decoded into \cspace, providing promising samples to a sampling-based planner (here, \rrtc is used).
We use the provided implementation of \cvae~\cite{Ichter2018}.
We test two variants of \cvae: ``\cvae w/o Task'', which has the same input as \sparkthree albeit with a fixed number of obstacles, and ``\cvae'', which includes the start and goal as recommended in~\cite{Ichter2018}. The latent space dimension was increased as proposed by~\cite{Kumar2019} for our high \dof problems. The same training data (500 paths) were given to \cvae as \spark which is approximately 
5000 training samples. 

All methods show marked improvement when the variance is low.
However, only \sparkthree, \flame, and \cvae retain performance as the axes of variation increases (\autoref{fig:small}b).
As environment variance grows, \thunder's sparse roadmap becomes less informative and falls back to planning from scratch.
Similarly, the performance of ``\cvae w/o Task'' degrades as diversity increases (possibly due to discontinuities).
\cvae greatly benefits from the addition of the start and goal, indicating it is highly informative for this task.
Increasing the size of training examples did not improve the performance of \cvae
Finally, note that \sparkthree and \flame have lower variance in solution time, showing both become more effective at solving the ``harder'' problems in the test set.

\subsection{Convergence and Scalability}
\label{subsec:convergence}
\begin{figure}
  \centering
	\includegraphics[width=0.90\linewidth]{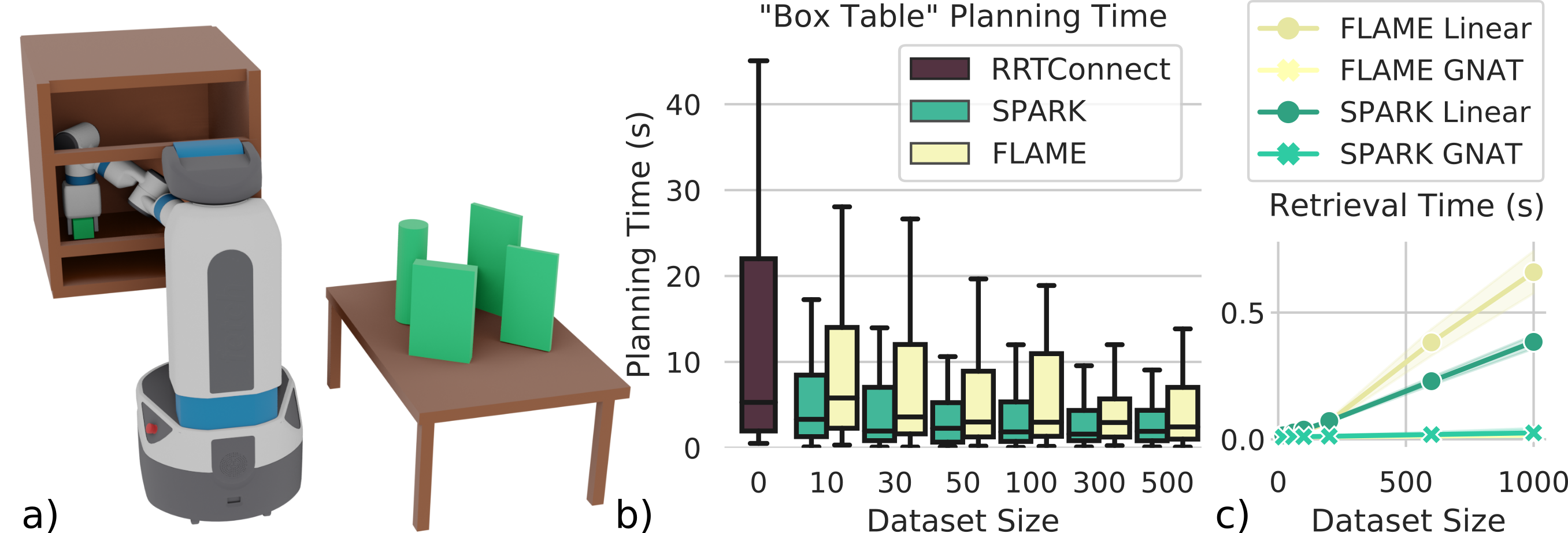}
  \caption{%
   \textbf{a)} The ``Box Table'' environment with a "place" task.
   The table and box vary relative to the robot in angular placement $\pm 30^\circ$ and position of X- and Y-coordinates by $\pm10$cm.
   \textbf{b)} Plot of average planning time (including retrieval) versus the size of the dataset, tested on 100 sampled environments with a timeout of $120$ seconds.
   \textbf{c)} Mean local primitive retrieval time versus the size of the dataset.
    %Both \sparkthree and \flame solve more problems faster than \rrtc.
     }
  \label{fig:table}

\end{figure}

\begin{figure}
  \centering
	\includegraphics[width=0.90\linewidth]{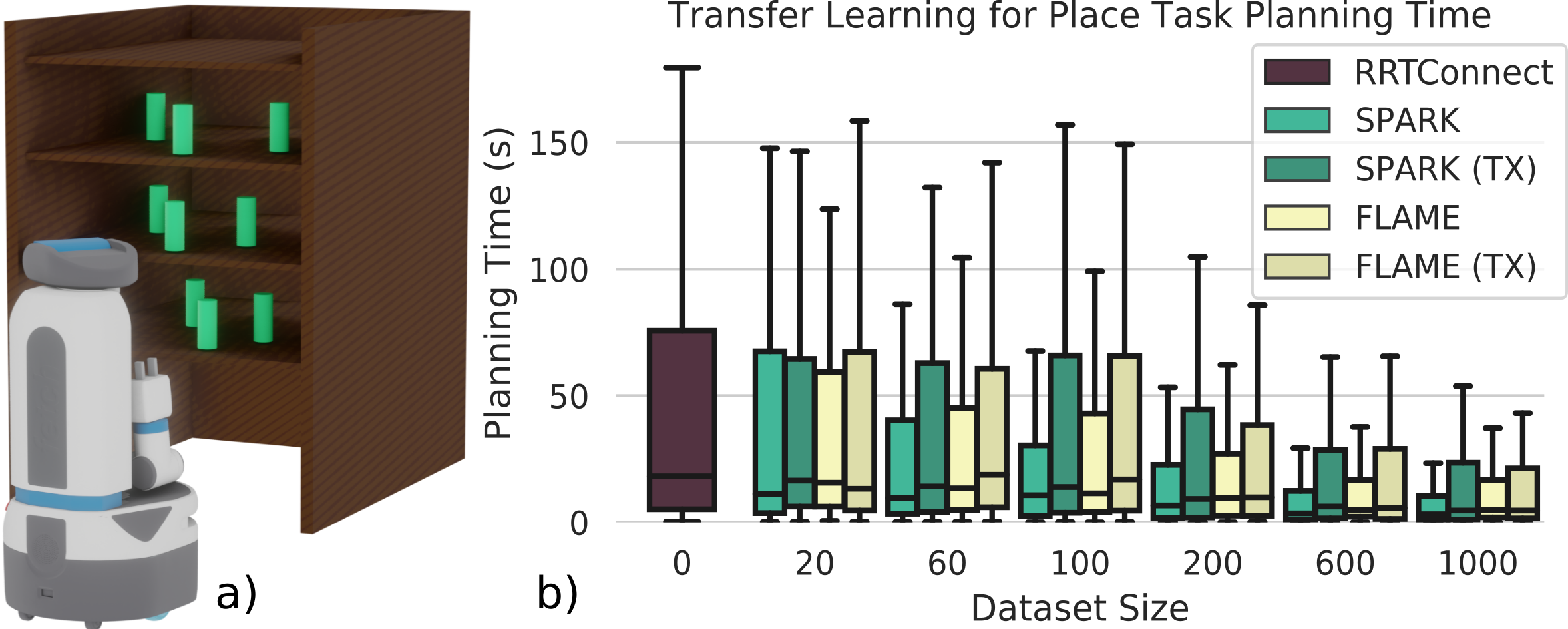}
  \caption{%
    \textbf{a)} The ``Large Shelf'' environment with a "pick" and "place" task.
    Three stacked shelves are generated similar to the ``Small Shelf'' environment (\autoref{fig:small}a).
    The stacked shelves vary relative to the robot in angular placement $\pm 90^\circ$ and position of X- and Y-coordinates by $\pm10$cm.
    %Starting from the Fetch's stow position, the deepest cylinder in a randomly selected shelf is used as the goal for a ``pick'' task.
    \textbf{b)} Plot of average planning time (including retrieval) versus the size of the dataset, tested on 200 sampled environments with a timeout of $300$ seconds.
    %The shelf varies in angular placement $\pm 90^\circ$ and position of X- and Y-coordinates by $\pm10$cm.
   % While more data is required than the ``Small Shelf'' environment, our approaches achieve better performance and solve more problems than uniform sampling.
   % Both \sparkthree and \flame solve more problems faster than \rrtc.
  }
  \label{fig:large}

  %\vspace{-1em}
\end{figure}

\autoref{fig:small}c, \autoref{fig:table}b and \autoref{fig:large}b present convergence results for \sparkthree and \flame for the ``pick'' task in the ``Small Shelf'' and the place task for ``Large Shelf,''``Box table'' environments.
\sparkthree and \flame both achieve significant improvement in average performance given only a few examples and quickly converge ($\approx 300$ for ``Small Shelf,'' `` Box Table'' and $\approx 600$ for ``Large Shelf''). 
\sparkthree has better asymptotic performance (especially in "Box Table") than \flame.
This is expected, as \flame uses the less informative but more general decomposition.
\autoref{fig:table}c shows the database retrieval time versus the size of the dataset using na\"ive linear search and \gnat-based search.
The number of local primitives for 1000 training examples is $\approx 12,000$ for \spark and $\approx 17,000$ for \flame.
\gnat scales much better than naive linear search.
%Linear search scales poorly with the size of the database, while \gnat scales well.
Although the retrieval overhead is negligible, we expect that it would affect performance if the database grows too large.

\subsection{Transfer}
\label{subsec:transfer}

\autoref{fig:large} presents transfer learning results for \sparkthree and \flame on a ``place'' task in the ``Large Shelf'' environment.
Here, we show how \sparkthree and \flame can learn from examples in one task and apply their experience to a different, harder task.
In \autoref{fig:large}, a baseline of \sparkthree and \flame trained and tested both on ``place'' tasks  (labeled as \sparkthree and \flame) is compared to \sparkthree and \flame trained on ``pick'' tasks and tested on the ``place'' task (labeled as \sparkthree (TX) and \flame (TX)).
Note that the ``place'' task changes the robot's geometry by attaching an object in the end-effector's grasp, making the planning problem harder.
Even when using experience gathered on a different, easier motion planning problem both \sparkthree and \flame demonstrate significant performance gains, successfully transferring their experience.
We hypothesize that \flame shows better performance in transfer learning as it uses an approximate decomposition, and thus can find more relevant experience than \sparkthree. 

\subsection{Real Robot}
\label{subsec:real}

\autoref{fig:real} shows timing results for \sparkthree and \flame applied to a real robot system.
To increase the statistical significance of our results, we use 6-fold cross-validation.
Additionally, we carefully tuned the range of \rrtc to obtain the best results possible for all methods (range $=2$ was used).
A \vicon camera system was used to gather object poses for \sparkthree.
The Fetch's depth camera was used to build an octomap~\cite{Hornung2013} for \flame.
Note that \flame can retrieve relevant experiences even from partial/noisy octomaps which is often the case for real systems. 
Both demonstrate significant improvement even with few given examples.

\section{Conclusion}
\label{sec:discussion}

We have presented two experience-based planning frameworks for arbitrary manipulators in \threed environments, instantiated in \sparkthree and \flame. 
\sparkthree uses an exact decomposition of the workspace into pairs of obstacles and achieves the best performance when this information is available.
\flame uses an approximate decomposition from a volumetric octree generated from sensor data and is comparable in performance with \sparkthree.
We demonstrated that, with only a handful of examples, both approaches confer significant improvement on planning time on a Fetch robot (8 \dof), on challenging manipulation tasks in environments that substantially vary, outperforming other learning-based approaches.
One limitation of our approaches is that due to our conservative metrics between local primitives, it is hard to generalize to environments that are far outside of the training distribution.
For example, the ``Large Shelf'' and ``Real Shelf'' share no local primitives (due to the height of the shelves) and thus do not transfer from one to the other.  
In the future, we will investigate more diverse environments and learning local primitive representations, metrics, and criticality tests.

\addtolength{\textheight}{-1.7cm} %use to equalize the references  

\bibliographystyle{IEEEtran}
\bibliography{references}

\end{document}